%% file: lrec2022-adr.tex
\documentclass[10pt, a4paper]{article}
\usepackage{lrec2022} 
\usepackage{multibib}
\newcites{languageresource}{Language Resources}
\usepackage{graphicx}
\usepackage{tikz}
\usepackage{tabularx}
\usepackage{soul}
\usepackage{booktabs}
\usepackage{lscape}
\usepackage{enumitem}   
	
\usepackage{color, colortbl}
\usepackage{xcolor}
	
\definecolor{lightcyan}{rgb}{0.88,1,1}

\definecolor{seabornBlue}{HTML}{0173B2}
\definecolor{seabornOrange}{HTML}{DE8F05}
\usepackage{titlesec}
\titleformat{\section}{\normalfont\large\bfseries\center}{\thesection.}{1em}{}
\titleformat{\subsection}{\normalfont\SmallTitleFont\bfseries\raggedright}{\thesubsection.}{1em}{}
\titleformat{\subsubsection}{\normalfont\normalsize\bfseries\raggedright}{\thesubsubsection.}{1em}{}
\renewcommand\thesection{\arabic{section}}
\renewcommand\thesubsection{\thesection.\arabic{subsection}}
\renewcommand\thesubsubsection{\thesubsection.\arabic{subsubsection}}

\usepackage{epstopdf}
\usepackage[utf8]{inputenc}

\usepackage[hidelinks]{hyperref}
\usepackage{xstring}

\usepackage{color}

\definecolor{mypink1}{rgb}{0.858, 0.188, 0.478}
\definecolor{titlecolor}{RGB}{74, 114, 159}
\definecolor{titledarkcolor}{RGB}{51,102,153}
\definecolor{LightGrey}{RGB}{232, 232, 232}
\definecolor{Grey}{RGB}{222, 223, 225}
\definecolor{DarkerGrey}{RGB}{215,217,219}
\definecolor{FontColor}{RGB}{131,136,138}
\definecolor{Red}{RGB}{204,0,0}
\definecolor{L-lig}{RGB}{25,124,192}
\definecolor{point-lig}{RGB}{54,104,163}
\definecolor{G-lig}{RGB}{62,66,68}
\definecolor{DarkRed}{RGB}{175,0,42}

\definecolor{Orange}{RGB}{240,163,10} 
\definecolor{Gray}{RGB}{186,200,211}
\definecolor{LightRed}{RGB}{214,98,93}
\definecolor{LightBlue}{RGB}{160,200,217}
\definecolor{LightGreen}{RGB}{130,161,119}
\definecolor{Violet}{RGB}{190,144,252}

\title{Cross-lingual Approaches for the Detection of Adverse Drug Reactions in German from a Patient's Perspective}

\name{Lisa Raithel$^{1,2,3}$, Philippe Thomas$^{1}$, Roland Roller$^{1}$, Oliver Sapina$^{1}$,\\
{\bf \large Sebastian Möller$^{1,2}$, Pierre Zweigenbaum$^{3}$}}

\address{$^{1}$Deutsches Forschungszentrum für Künstliche Intelligenz (DFKI) Berlin, 10559 Berlin, Germany\\
$^{2}$Technische Universität Berlin, 10623 Berlin, Germany\\
$^{3}$Université Paris-Saclay, CNRS, Laboratoire interdisciplinaire des sciences du numérique (LISN),\\ 91405 Orsay, France\\
 lisa.raithel@dfki.de
 }

\abstract{
In this work, we present the first corpus for German Adverse Drug Reaction (ADR) detection in patient-generated content. 
The data consists of 4,169 binary annotated documents from a German patient forum, where users talk about health issues and get advice from medical doctors.
As is common in social media data in this domain, the class labels of the corpus are very imbalanced. 
This and a high topic imbalance make it a very challenging dataset, since often, the same symptom can have several causes and is not always related to a medication intake.
We aim to encourage further multi-lingual efforts in the domain of ADR detection and provide preliminary experiments for binary classification using different methods of zero- and few-shot learning based on a multi-lingual model.
When fine-tuning XLM-RoBERTa first on English patient forum data and then on the new German data, we achieve an F1-score of 37.52 for the positive class.
We make the dataset and models publicly available for the community.
 \\ \newline  }

\begin{document}

\maketitleabstract

\section{Introduction}

Adverse drug reactions (ADRs) are a major and increasing public health problem and exist all over the world, mostly under-reported \cite{alatawi_empirical_2017}.
They describe an unanticipated, negative reaction to a medication.
Of course, new drugs are tested extensively when being developed, but certain vulnerable groups, like pregnant or elderly people, are rarely part of clinical trials and still might be in need of medication at some point.
Furthermore, clinical trials and also physicians prescribing medications cannot cover every potential use case.
Therefore, an efficient monitoring of ADRs from different angles is important and necessary for improving public health and safety in medication intake.

One of those angles is the classification of documents containing mentions of ADRs.
While there exist various resources for training NLP models on the task of ADR classification, for example scientific publications or drug leaflets, the world wide web provides a more up-to-date and diverse source of information, especially social media and patient fora \cite{sarker_portable_2015}.
Here, patients freely write about their experiences during medication therapy. 
By using their own non-expert language, e.g.~using laymen terms for describing their situation, they create a rich data source for pharmacovigilance from a patient's perspective in a wide variety of languages.
Therefore, there is theoretically a huge amount of text to leverage. On the other side, however, this text is noisy (e.g.~spelling mistakes), incomplete (e.g.~deleted messages) and  the use of laymen terms and abbreviations (e.g.~``AD'' as a generic name for any anti-depressant) complicates the processing of these data \citelanguageresource{seiffe_witchs_2020-1,basaldella_cometa_2020}.
It is furthermore difficult to collect these resources (e.g.~finding informative keywords to search Twitter) and researchers are soon confronted with privacy issues when handling social media data.
Thus, only a small amount of (annotated) data is publicly available for research and unfortunately, most of these data are in English.

Another issue arises when looking at the label distributions of the described resources.
Depending on where the dataset originates from and how it was collected, the distribution of labels for text classification is skewed in either direction: for instance, data from the CADEC corpus \citelanguageresource{karimi_cadec_2015} tend to include a lot more positive documents (that is, documents containing adverse drug reactions) than negative ones, while for example HLP-ADE \cite{magge_deepademiner_2021-1}, a corpus of tweets, contains 13 times more negative than positive examples.

Thus, before working on the more specific task of extracting adverse effects and their corresponding drugs, and also the relations between them, there needs to be a reliable way to distinguish texts mentioning ADRs from those without ADRs.
This first step is still necessary even in the era of deep learning \cite{magge_deepademiner_2021-1}.
Therefore, we manually annotated a dataset of German health-related user posts and employed it for zero- and few-shot experiments using a Language Model (LM) equipped with a binary classification head.
With this, we want to investigate whether it is possible to use an LM fine-tuned on one language (the source language) to classify documents from another language (the target language) by using only a small number of available documents in the ADR domain. 
We do this in a ``true'' few-shot scenario \cite{kann_towards_2019-1,perez_true_2021}, assuming that we do not only have a small number of shots for the fine-tuning training set, but also only a small number of shots for the development and test set.
We focus our efforts on finding documents containing ADRs, i.e.~the positive class (label 1), since these are the most interesting to us.
Our contributions are as follows:

\begin{description}
    \item[Dataset] We provide a binary annotated corpus of German documents containing adverse drug reactions. 
    The dataset is challenging, since its class and topic distributions are imbalanced, and the texts are written in everyday language.
    \item[Few-Shot Classification] We experiment with different model settings and data combinations to transfer knowledge between English and German.
    In several few-shot settings, we come close to the performance of fine-tuning on the full German (training) data while achieving a higher recall, making the few-shot models a potentially better filter for unlabeled data.

    \item[Error Analysis] We conduct an error analysis on the predicted results to provide future directions of research for handling imbalanced, small and user-generated data. 
\end{description}

Models, code and dataset can be found on github\footnote{\url{https://github.com/DFKI-NLP/cross-ling-adr}. The dataset will only be accessible via a data protection agreement.}.

\section{Related Work}

Before social media became popular, the detection and extraction of ADRs was mostly conducted on electronic health records (EHRs) and clinical reports \cite{uzuner_2010_2011,lo_mining_2013,sarker_portable_2015}. 
Nowadays, however, not all drug effects are reported to healthcare professionals, but are also widely discussed online.
This, and the rise of deep learning, spurred the collection of datasets (mostly in English) and the introduction of shared tasks and challenges, such as the SMM4H series \cite{weissenbacher_overview_2018-1,weissenbacher_overview_2019,klein_overview_2020}.
The methods of choice for tackling these tasks often included rule-based approaches and ensembles of statistical classifiers, e.g.~Support Vector Machines based on static word embeddings \cite{sarker_portable_2015,nikfarjam_pharmacovigilance_2015}.
Now, the majority of approaches uses deep neural nets \cite{minard_irisa_2018,wunnava_adverse_2019} and with the introduction of transformer models \cite{vaswani_attention_2017-1}, especially BERT \cite{devlin_bert_2019-1} and all its variants are taking over.

For example, \newcite{chen_hitsz-icrc_2019} use BERT combined with a knowledge-based approach to achieve first place in the SMM4H 2019 task with an F1-score of 62.89 for ADR classification. 
In SMM4H 2020 Task 2, the best system employed a RoBERTa model and achieved an F1-score of 64.0 for the positive class \cite{wang_islab_2020}.

Also, joint learning approaches are getting more attention.
\texttt{DeepADEMiner} \cite{magge_deepademiner_2021-1}, for instance, is a pipeline for classifying, extracting and normalizing ADRs from Twitter data in one go.
They also publish a naturally distributed dataset of tweets (7\% positive documents) on which they achieve an F1-score of 63.0 for the task of document classification using a RoBERTa model \cite{liu_roberta_2019}.

\newcite{raval_exploring_2021-1}, on the other hand, employ a T5 model \cite{raffel_exploring_2020} to jointly learn ADR classification and extraction of ADRs, drugs and drug dosages as a sequence-to-sequence problem. 
Their proposed method applies the original T5 task balancing strategies, but adds dataset balancing to account for different dataset sizes, domains and label distributions.
On CADEC \citelanguageresource{karimi_cadec_2015} and the SMM4H 2018 tasks 1 and 2 \cite{weissenbacher_overview_2018-1}, the authors achieve an F1-score for the positive class of 98.7, 69.4 and 89.4 respectively (for ADR detection).
Finally, the authors also apply their model on the very imbalanced French SMM4H 2020 dataset \cite{klein_overview_2020} in a zero-shot fashion and achieve an F1-score of 20.0.

The latter is one of the few approaches to tackle ADR data that is not in English.
Just recently, however, we can see an increased effort to publish non-English data for the detection of ADRs which we will describe in the next section.
\begin{table*}[h]
\centering
\footnotesize
\centering
\begin{tabular}{p{.43\linewidth}p{.43\linewidth}r}
\multicolumn{1}{c}{\textbf{German}} &
  \multicolumn{1}{c}{\textbf{English Translation}} &
  \multicolumn{1}{c}{\textbf{Label}} \\ \midrule
Ach ja, vielleicht für einige noch interessant.  Hatte in meiner Verzweiflung ja auch ein \colorbox{seabornOrange}{AD} ausprobiert. Bei mir hat es den \colorbox{seabornBlue}{\textcolor{white}{TSH voll hochgetrieben}}! Davon hatte der Psychiater  noch nie gehört...echt verrückt. Geholfen hat  es übrigens überhaupt gar nicht. Konnte es  zum Glück problemlos ausschleichen. &
By the way, this might be interesting for some.  In my despair, I had also tried an \colorbox{seabornOrange}{AD}. It \colorbox{seabornBlue}{\textcolor{white}{drove my TSH all the way up}}! The psychiatrist had never heard of this...really crazy. By the way,  it didn't help at all. Fortunately, I was able to  phase it out without any problems.
& 1
\\
\midrule
Hallo. Man muss sich nur mal eine Zigarettenschachtel ansehen.  Die Warnhinweise sind nicht umsonst  drauf. Ich bin seit Ewigkeiten Raucherin  mit 5 jähriger Unterbrechung, aber ich  Depp habe wieder angefangen. Es ging  mir 100 Mal besser ohne Glimmstengel.  Seit ich in den WJ bin, wird mir nach der  ersten Kippe am Tag schummelig. Das  sagt wohl alles. Ich verteufel meine Sucht. Lg &
Hi. You only have to look at a packet of  cigarettes. The warnings are not there for  nothing. I've been a smoker for ages with a  5 year break, but I started again. I felt a 100  times better without the cigarettes. Since I've  been in MP, I get woozy after the first smoke  of the day. I guess that says it all. I'm  demonizing my addiction. Cheers
& 0
\\
\bottomrule
\end{tabular}
\caption{A positive (top) and a negative example (bottom) from the \textit{Lifeline} corpus. A positive document is one containing an adverse effect of a drug. \colorbox{seabornOrange}{Medication} and \colorbox{seabornBlue}{\textcolor{white}{adverse effect}} for the positive sample are color-coded. Note the use of (very general) abbreviations (AD = anti-depressant, TSH = Thyroid-stimulating hormone, MP = meno pause) and the descriptions in colloquial speech.}
\label{tab:example}
\end{table*}
\section{Datasets}

We first describe the new German dataset we provide and then shortly summarize other available, non-German datasets.

\subsection{The \textit{Lifeline} corpus}

For populating our corpus, we choose the German \textit{Lifeline} forum\footnote{\url{https://fragen.lifeline.de/forum/}}, a forum where people write about health issues, but also about other topics.
Users can only participate in the discussion if they are registered, but all questions and answers are freely accessible without registration.
The forum is anonymous as well.
With permission of the forum administrators, we downloaded the existing user threads\footnote{Data downloaded in July 2021.} and gave the accordingly prepared documents to one of our annotators, a final year student in pharmacy with some practical experience in the handling of medications.

\begin{table}[ht]
\centering
\footnotesize
\centering
\begin{tabular}{lrr}
\multicolumn{1}{c}{\textbf{topic}} & \multicolumn{1}{c}{\textbf{train/dev}} & \multicolumn{1}{c}{\textbf{test}} \\ \midrule
{women's health}    & {2541}              & {634}          \\
{cosmetic OPs}      & {166}               & {47}           \\
{skin}              & {129}               & {36}           \\
{bones}             & {125}               & {29}           \\
{gen. med.}         & {117}               & {20}           \\
{heart}             & {92}                & {26}           \\
{nerves}            & {44}                & {11}           \\
{men's health}      & {22}                & {3}            \\
{sports}            & {21}                & {6}            \\
{infections}        & {21}                & {5}            \\
{nutrition}         & {19}                & {9}            \\
{int. organs}       & {15}                & {3}            \\
{allergies}         & {8}                 & {2}            \\
{life}              & {7}                 & {2}            \\
{gastroint. system} & {7}                 & {2}            \\
{\textbf{avg \#tokens}}    & {110.6} & {109.9} \\
{\textbf{avg \#sentences}} & {8.3}   & {8.2}   \\ \bottomrule
\end{tabular}
\caption{The distribution of topics for train/dev and test set in the \textit{Lifeline} corpus. 
Bottom: the average number of tokens and sentences per document.}
\label{tab:topics}
\end{table}

\begin{table*}[h]
\centering
\begin{tabular}{@{}lrrrrllr@{}}
\toprule
\textbf{lang.} &
  \textbf{overall} &
  \textbf{neg} &
  \textbf{pos} &
  \textbf{ratio} &
  \textbf{type} &
  \textbf{annotation} &
  \textbf{authors} \\ \midrule
es & 400  & 235  & 165 & 1.4 : 2 & forum        & entities    & \citelanguageresource{segura-bedmar_detecting_2014} \\
fr  & 3033 & 2984 & 49  & 61 : 1  & Twitter      & binary      & \citelanguageresource{klein_overview_2020}          \\
ru & 370  & -    & -   & -       & drug reviews & multi-label & \citelanguageresource{alimova_machine_2017}         \\
ru &
  500 &
  - &
  - &
  - &
  drug reviews &
  multi-label + entities &
  \citelanguageresource{tutubalina_russian_2021-1} \\
ru & 9515 & 8683 & 842 & 10 : 1  & Twitter      & binary      & \citelanguageresource{klein_overview_2020}          \\
ja+en &
  169 &
  - &
  - &
  - &
  forum &
  entities &
  \citelanguageresource{arase_annotation_2020} \\
de  & 4169 & 4068 & 101 & 40 : 1  & forum        & binary      & ours                                                \\ \bottomrule
\end{tabular}
\caption{Other non-English social media corpora for the detection (and partially extraction) of ADRs. en=English, fr=French, ru=Russian, ja=Japanese, de=German. Note the label imbalance in all datasets.}
\label{tab:data}
\end{table*}

Note that one document corresponds to a complete forum post and thus (usually) contains more than one sentence.
This, of course, comes with a caveat: sometimes, documents contain mentions of adverse reactions to one drug, but also positive reactions to another drug, leaving the final class label unclear.  
However, to facilitate annotation, we kept the guidelines very simple: each document containing at least one adverse reaction was to be labeled as positive (Class~1), all others were to be labeled as negative (Class~0). 
The only exceptions were documents that contained speculations in which people only knew about side effects from hearsay or rumors and did not experience them themselves. 
Those samples were flagged by the annotator and removed for future investigation.
After one round of training and discussing the annotations, our annotator labeled 4,169 documents (flagged ones already excluded) using the annotation tool Prodigy\footnote{\url{https://prodi.gy/docs}}.
The average progress was 100 documents in about one hour.

\begin{figure}[h]
    \centering
    \includegraphics[scale=0.25]{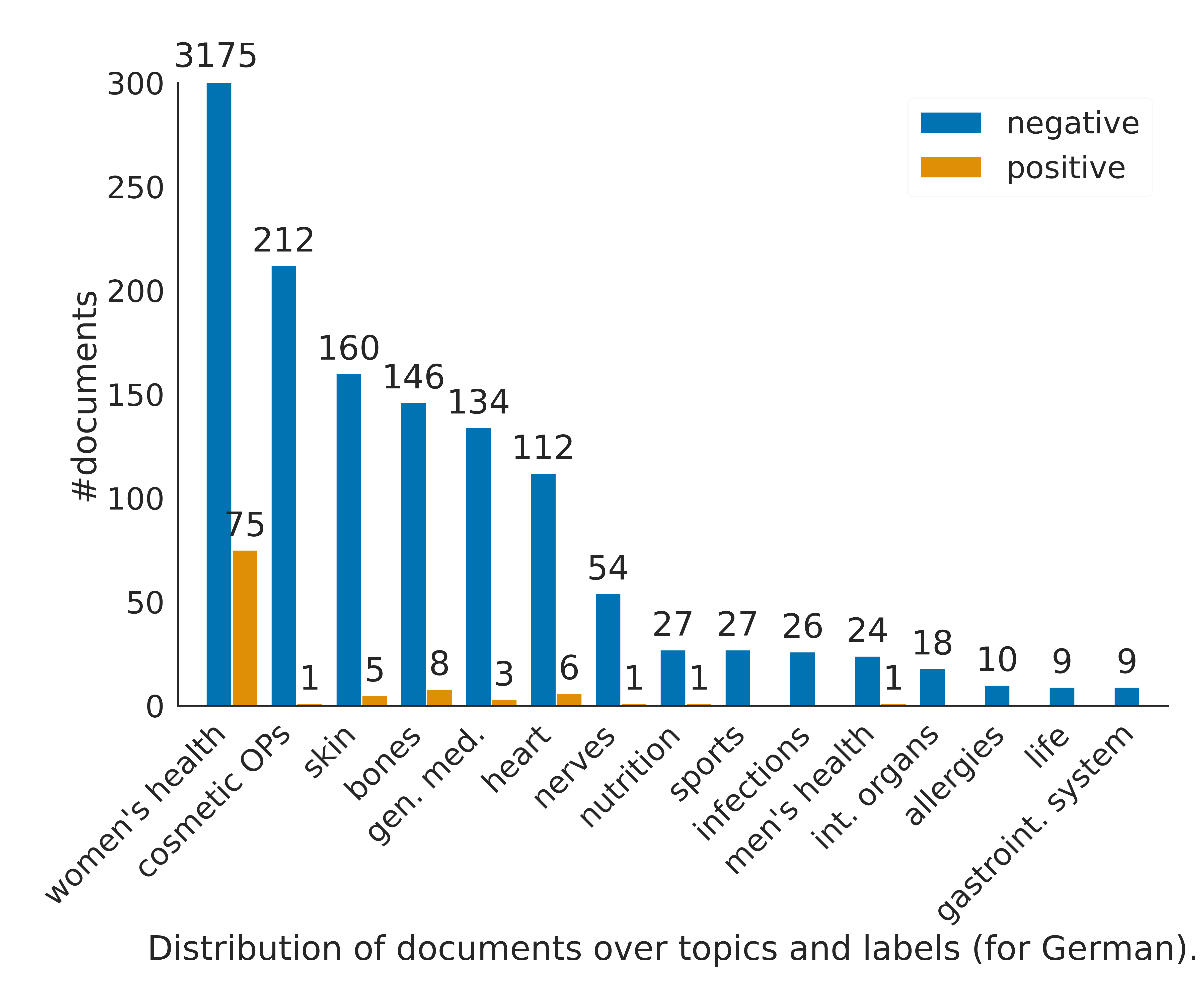}
    \caption{The distribution of topics in the new German corpus. The blue bar represents the negative documents, the orange bar the positive ones. Note the huge difference in the number of documents per topic especially in the \emph{women's health} thread.}
    \label{fig:topics}
\end{figure}

The resulting \textit{Lifeline} dataset for binary classification of ADRs contains 101 positive and 4,068 negative examples (positive to negative ratio $\sim 1:40$).
In Table~\ref{tab:example} we show one positive and one negative document as an example.
Figure \ref{fig:topics} shows the distribution of topics over the entire dataset.
Note the huge amount of documents in \emph{women's health} (3,175 documents) compared to the other topics.
In Table~\ref{tab:topics}, we show the distribution of topics divided into  train/dev and test set.
The average number of tokens per document is approximately 110.6 for the train/dev set and 109.9 for the test set.
Further, the documents contain about 8.3 sentences in the train/dev set and 8.2 sentences in the test set.

\subsection{Other available data}

Apart from the already mentioned corpora, other (English) datasets might contain for instance tweets \cite{magge_deepademiner_2021-1}, case reports \citelanguageresource{yada_towards_2020} or the content of PubMed abstracts \citelanguageresource{gurulingappa_development_2012}.
For our experiments, we combine the datasets CADEC \citelanguageresource{karimi_cadec_2015} and PsyTAR \citelanguageresource{zolnoori_psytar_2019}, since both corpora are based on a patient forum\footnote{\url{www.askapatient.com}}.
We used only those two, because we assumed data from another domain, e.g.~Twitter or more structured sources, might complicate the transfer between the English and the German data.
Together, they comprise 2,137 documents, with 1,683 positive and 454 negative examples.

\begin{figure}
    \centering
    \includegraphics[scale=0.5]{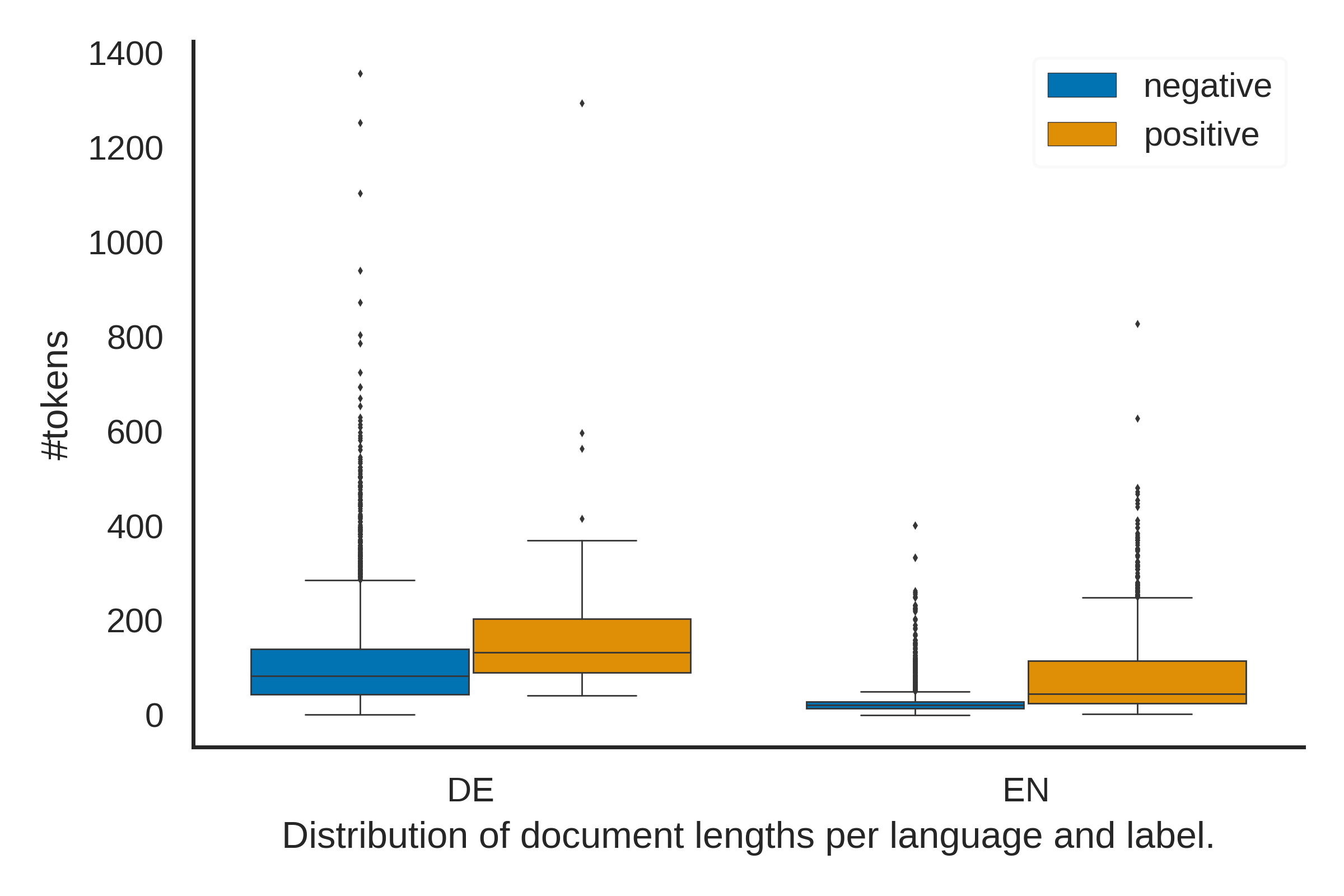}
    \caption{The number of tokens per label and per corpus (not divided by train/dev and test). ``DE'' corresponds to the \textit{Lifeline} corpus, while ``EN'' corresponds to the combination of CADEC and PsyTAR. For both corpora, the documents containing no ADRs tend to be shorter.}
    \label{fig:label_vs_length}
\end{figure}

Finally, we show in Table \ref{tab:data} the non-English social media datasets that are (at least partially\footnote{For the French and Russian Twitter datasets, the test sets are unfortunately not available, they were part of the SMM4H 2020 shared task.}) publicly available, including our new German corpus.
Three of those comprise data from patient fora (Spanish, Japanese, German), two are created from Twitter messages (French, Russian) and two are collected drug reviews (Russian). 
Note the strong label imbalance for all corpora, and the reverse labels distribution for the Spanish data (and also for the English data we use).
Moreover, the corpora vary strongly in their overall sizes and length of documents: for example, Twitter messages are usually rather short, while forum posts might contain several hundred tokens.
As we can see in Figure \ref{fig:label_vs_length} comparing the German and the English dataset we use for our experiments, the length of a user post also varies with its content: documents containing ADRs tend to be longer than those without ADRs.

Since English and German are typologically closer than English and the other languages listed in Table \ref{tab:data}, we focus the experiments on the transfer of the English knowledge (\emph{source language data}) to the German data (\emph{target language data}).
The target language data contains the German forum posts as described above.
Note the reverse class label distribution for both datasets.

\section{Experiments}

In the following, we describe the experimental setting and models we used.
Details for the specific fine-tuning (hyper-) parameters can be found in the appendix.
Our goal is to classify target language documents into those containing ADRs (the positive Class~1) and those that do not contain any ADR (the negative Class~0).

As baseline, we use a Support Vector Machine (SVM, \cite{boser_training_1992}) and for the neural network approach, we chose a transformer model \cite{vaswani_attention_2017-1}, since these are capable of incorporating contextual knowledge to a certain degree.
Since we only have a small number of positive samples in the target data, we first fine-tune a model on the source data and then try to transfer the learned knowledge via (i) zero-shot ``learning'', (ii) a second fine-tuning on \emph{all} the target language training data (henceforth called \emph{full model}), and (iii) few-shot learning.
For (iii), we experiment with different ``modes'': per-class few-shot learning (per\_class), adding negative examples to the shots (add\_neg), and adding negative as well as source language examples to the shots (add\_source). 
The modes are explained in more detail in Section \ref{subsec:fine_tuning}.

\subsection{Pre-processing}

Pre-processing is handled in a very simple way: Before feeding the documents to the model, we replace URLs, user names, dates, and similar occurrences with generic names, e.g.~\texttt{<URL>}, using ekphrasis \cite{baziotis_datastories_2017}.
For the baseline models, the documents are then tokenized simply by white space and for the transformer models, tokenization is done by the wordpiece tokenizer \cite{wu_googles_2016-1} of the respective model.
Documents with less than four tokens are filtered out, those that are longer than 300 tokens are truncated. 
This setup achieved the best results during preliminary experiments.
Both datasets are divided into a training/development (train/dev, 80\%) and a test set (20\%) via a stratified split corresponding to the distribution of labels.

\subsection{Baselines}

We train a Support Vector Machine on the target language data using fasttext embeddings \cite{bojanowski_enriching_2017} and sklearn \cite{pedregosa_scikit-learn_2011}.
The embeddings for one document are calculated as the average over the word vectors. 
The models are trained with class weights (``balanced'') and otherwise default parameters, and tested on the same (German) test set on which the transformer models are tested.
For comparison with the neural approach, we use the available shot data, add negatives to the shots and also add negatives plus source language data to the shots, using aligned embeddings \cite{joulin_loss_2018}.
Further, we train an SVM on the entire target training data (full SVM).

\subsection{Two-Stage Fine-Tuning}\label{subsec:fine_tuning}

A perfect transformer model to start with would be a multi-lingual model trained on \emph{health-related}, \emph{user-generated} texts.
However, since this rare combination does not exist (yet), we explore XLM-RoBERTa \cite{conneau_unsupervised_2020}, a multi-lingual model trained on general domain texts (henceforth XLM-R), and BioRedditBERT \cite{basaldella_bioreddit_2019-1}, a biomedical (English) model fine-tuned on user posts from Reddit\footnote{\url{https://www.reddit.com/}} (henceforth BRB).

First, for both model types, we conduct a hyper-parameter search to find the best parameters for our task on the source language data with respect to macro F1-score on the development set.
For this, we use the Weights \& Biases sweeps framework \cite{biewald_experiment_2020}. 

Using the determined hyper-parameters and ten different seeds for model initialisation, we fine-tune (fine-tuning 1) ten source language models on the \emph{source data} to account for the instability of language models \cite{devlin_bert_2019-1}: XLM-R$_{1}$ -- XLM-R$_{10}$ (the same for the BRB model).
 
We hypothesise that using all target language data will bias the model towards the negative class (the negative to positive ratio is 40 : 1) and want to counter-act this by testing several few-shot scenarios.
Here, however, we want to apply a ``true'' few-shot setting \cite{perez_true_2021} and thus do not use an extensive development set to optimize the models on.
Thus, the dev set has always the exact same number of examples and classes as the train set.
We evaluate the following few-shot scenarios:

\begin{description}
    \item[per\_class] We use the exact same number of documents for both labels in train and dev set. For example, if we use \emph{ten shots}, we have five positive and five negative examples in both the train and dev set.
    
    \item[add\_neg] We refer with the term ``few-shot'' only to the positive examples and add a certain amount of negative samples to the train and dev set. Using the example of \emph{ten shots} again, we construct a train set of ten positive and \{100, 200, 300, 400\} negative examples. The same goes for the dev set.
    This approach serves to approximate the ``natural'' label distribution of the target data.
    
    \item[add\_source] We again refer only to the positive examples when using the term ``few-shot'', add \{100, 200, 300, 400\} negative examples and \emph{additionally} \{100, 200, 300, 400\} random samples from the source language for both train and dev set.
    We assume that this approach might help to reduce the \emph{catastrophic forgetting} \cite[i.a.]{MCCLOSKEY1989109} of language models.
\end{description}

Note that we only use shots of 10 and 40 to reduce the amount of experiments and since there are only 101 positive samples in the train/dev set of the target language in total, we can only experiment with up to 40 shots. 
The corresponding test set then contains 21 positive examples.
We then proceed as follows:

\begin{enumerate}
    \item We choose five seeds for sampling from the \emph{target train/dev set}, creating five different train/dev sets to sample the shots from. 
    
    \item We freeze all \emph{source language} models except for their classifier and fine-tune (fine-tuning 2) on the five sampled sets with the few-shot approaches described above. This results in \{XLM-R$_{fine1}$, ..., XLM-R$_{fine10}$\} for every seed and every scenario (again, the same goes for the BRB models). 
     
    \item Each model in each scenario is applied to the fixed target language test set; the final prediction is decided by majority voting of all ten models. 
    
    \item Finally, the performance per scenario is averaged over the five seeds.
\end{enumerate}

The experimental setup is visualized in Figure~\ref{fig:exp_setup} in the Appendix.
To compare the few-shot scenarios, we also fine-tune the ten XLM-R models (\textsc{XLM-R}$_{1}$, ..., \textsc{XLM-R}$_{10}$) on \emph{all} available target language training data, called the \emph{full model} (XLM-R$_{full}$), and finally also apply the \emph{not} fine-tuned \textsc{XLM-R} and \textsc{BRB} models in a zero-shot fashion to the target language.

Finally, we also try boosting the performance via rule-based post-processing, since we observed  that many false positives discuss health issues that are similar to adverse reactions.
After calculating the voting winners, we use an extensive medication list\footnote{22,827 medication names copy-pasted from a German information website about health-related topics (\url{https://www.apotheken-umschau.de/medikamente/arzneimittellisten/medikamente_a.html})} and a self-created shorter list related to women's health topics (the biggest topic in the dataset) and abbreviations to check each document's predicted label.
If it is positive but \emph{does not} include a drug name from the medication list, we switch the label to negative.
Conversely, if the label is positive and \emph{does} include a word from the women's health list, we switch the label to negative as well.
Both checks are performed independently and we calculate the final scores for each approach.

\input{results_table}

\section{Results}

We now present the results, ordered by language and fine-tuning approach.

\subsection{Source Language (English)}

The results for the first round of fine-tuning, i.e.~fine-tuning XLM-R and BRB on the source language data, can be found in Table \ref{tab:english_results_xlm} and \ref{tab:english_results_brb} in the Appendix.
For both models, we can see a clear tendency to perform better for the majority class---in this case it is the positive one.
XLM-R achieves an average F1-score of 91.03 ($\pm$0.67), while for the negative class, the average F1 is 65.20 ($\pm$2.81).

\subsection{Target Language (German)}

The results on the target language can be found in Table \ref{tab:final_results}: The second block (rows 6 and 7) shows the zero-shot approach, the third block (rows 8 - 13) shows the results of XLM-R and the last block (rows 14 - 17) shows the performance of BRB.
Note that we only show the best performance(s) with respect to F1-score of the positive class for every setting and leave out classifiers that did not predict any positives.

\paragraph{Baseline} The SVM baseline achieves a performance of 17.39 F1 for the positive class when we train the model on \emph{all available} German training data.
This performance is comparable to the \emph{few-shot} performance of the transformer models.
When using shot data, the F1-score for the positive class decreases by at least 4 points, depending on the applied few-shot method.
However, the SVM model combined with the per\_class strategy and 40 shots achieves the highest recall for the positive class overall.
The results of the baselines can be found in the first block of Table \ref{tab:final_results}.

\paragraph{Zero-Shot}
The zero-shot approach mainly stands out because of the second highest recall (95.23) and third-best AUC score for the positive class achieved by XLM-R (see the second block in Table \ref{tab:final_results}).
In contrast to that, the zero-shot approach using BRB is much more biased towards the negative class. 

\paragraph{Full fine-tuning}
Against our assumptions, a second fine-tuning of the already fine-tuned XLM-R$_{\mathit{full}}$ with the \emph{full target training dataset} achieves the highest F1-score for the positive class overall (37.52 $\pm$6.65) (Table \ref{tab:final_results}, third block, first row).
It also achieves the highest precision for the positive class, as well as the highest recall for the negative class.
Note, however, the rather low AUC score (64.0 $\pm$3.68) and the high standard deviations for both precision and recall of Class~1.
Also, the recall for Class~1 (28.57 $\pm$7.53) is amongst the lowest for all experiments.

\paragraph{per\_class}
In the per-class scenario, we can see that the 40-shot approach achieves a slightly higher F1-score for the positive class for both models.
This is somewhat surprising, we would have expected a higher difference in this scenario because of the additional positive samples during fine-tuning.
Note the high AUC for the 40-shot setting of XLM-R.
Otherwise, this approach ranges in the lower area with respect to F1-score for Class~1.

\paragraph{add\_neg}
Adding negatives during fine-tuning does not help in the 10-shot setting: here, for both models, we do not find any positives (and therefore, the results are not presented in the table).
For 40 shots, we find that adding 100 random negative samples works best.
Here, XLM-R clearly wins over BRB.

\paragraph{add\_source}
Adding negative target \emph{and} random source language data achieves (with 40 shots) the second best results after the full training. 
For XLM-R this works best for 10 shots when adding 100 negative and 200 source language examples, while the 40-shot scenario is improved by 300 negative and 300 source language examples.
Moreover, this setting achieves the best AUC score overall (78.95 $\pm$2.67 for 10 shots), even compared to the full-data model, and has a lower standard deviation than when only adding negatives. 
BRB does not find any positive examples with only 10 shots, but when adding 100 negative and 200 source language examples, it comes close to XLM-R's performance.
Also, the described post-processing can improve the F1-scores for the positive class: 
XLM-R increases its F1-score from 15.03 ($\pm$1.26) to 21.22 ($\pm$1.03) for 10 shots and from 22.55 ($\pm$3.42) to 28.56 ($\pm$2.29) for 40 shots;
BRB increases its performance from 22.23 ($\pm$4.06) to 25.33 ($\pm$4.29).

\subsection{Error Analysis}

We now provide an error analysis of the best performing model XLM-R$_{\mathit{full}}$ with respect to the falsely predicted documents.
Out of the 824 documents (21 positives, 803 negatives), the best model predicted 8/21 positives  and 796/803 negatives correctly, leaving 20 documents predicted incorrectly.

For the 13 false negatives, we can find no clear indication why those were missed except for one example where the document was cut off before the user was talking about side effects.
We notice some spelling mistakes and unclear formulations but nothing ``human-unreadable''.
Some adverse reactions are mentioned implicitly, though, or only very briefly.
On the other side, some of the false negatives are very clearly describing the problems, even mentioning the word ``side effects'', thus it is not obvious to us why the document was classified as negative.
One document describes the reactions partially in a positive light (weight gain), this sentiment might also mislead a model that is biased towards more negative sentiments with respect to adverse drug effects. 

Regarding the 7 false positives, we find several examples of persons talking about side effects they experienced \emph{before} taking the new drug the post is about.
Also, we find posts describing health issues that can be easily confused with side effects, and also one occurrence where the reactions came from \emph{not} taking the drug.

Against our expectations, we cannot see problems in the predictions with respect to the topic distribution, i.e.~there is no clear bias of the models towards performing better on documents from the women's health topic.
For most documents it is not clear why they were not correctly classified.
Here, a larger test set would probably help in the analysis.

\section{Discussion}

We find that, unsurprisingly, label imbalance (more positives than negatives in the first round of fine-tuning, vice-versa in the second round of fine-tuning) has the strongest influence on performance.
This is evident in both the baselines and the neural approaches.
It is interesting, however, that the models seem to perform better when having \emph{more, but imbalanced} data than when having carefully balanced data, as is the case for the per-class setting.
This might also be interrelated with the small number of samples overall.

Reminding the models of the original data (in our case the source data) gave the best results apart from the full data model, confirming again the phenomenon of catastrophic forgetting.
Adding source examples to the full data model training might therefore also improve the performance in the full data scenario.

Compared to almost all other settings, the full data model has a very low recall for the positive class (28.57 $\pm$7.53) as well as one of the lowest AUC scores.
This is both interesting and unfortunate, since it probably cannot be used as a filter for subsequent tasks, such as ADR span identification or ADR-drug relation extraction. 

Further, we conclude that the multi-lingual aspect of XLM-R seems to be of more importance than the user-data content of BRB, since in most cases XLM-R outperforms BRB.
However, it is still interesting to see that even BRB, pre-trained purely on English data, can produce some results on the German data, coming close to the performance of XLM-R in the add\_source setting.

We can also see a very high instability in the models, coming from the Language Models themselves but also from the sampled data (influenced by the seeds). 
Thus, it might help to carefully select the samples we use for few-shot training in case we have enough to choose from. 
However, we do not know if a model prefers the same examples as we humans do to learn better.

Further, one of the bigger issues with the presented models and dataset is that the model is often not able to distinguish side effects induced by drugs from accompanying effects of menopause (or other health-related issues).
We tried to counteract this by applying the described post-processing, and it helped, at least for the add\_source scenario, but not enough.
Admittedly, this might be improved by a better medication list: our list contains mostly the original drug names, and those are often not used by patients.

Of course, since German is very close to English, it remains to be seen whether the presented scenarios are also applicable to more distant languages.
There might also be some cultural aspects in the handling of adverse effects in online user forums for the different languages.

\section{Conclusion}

We have presented the first German corpus for the detection of adverse drug reactions from a patient's perspective, that is, a corpus created from user posts in a health-related online forum.
Further, we described a series of experiments to find the documents containing adverse drug reactions using a multi-lingual approach and comparing it with few-shot scenarios.
We experimented with a user-centred model and a more general multi-lingual model and found that classification performance benefits from the multi-lingual aspect.

The classifiers with a high recall can still be used as a filter for improving the downstream performance for ADR recognition and ADR-drug relation extraction, even though their F1-scores are not the best ones. 
Furthermore, it might be interesting to try character-based models, like e.g.~CharacterBERT \cite{el_boukkouri_characterbert_2020}. 
Although this is not a multi-lingual model it might perform better on rare words like medication names, and it might handle spelling mistakes better than wordpiece-based models.
Including negation of ADRs \cite{scaboro_nade_2021-1} and other corpora, e.g.~the TLC corpus \citelanguageresource{seiffe_witchs_2020-1}, to disambiguate user terms using a mapping from technical to laymen terms and vice versa might also be beneficial for the performance.

\section{Acknowledgements}

This research was funded by the Agence Nationale de la Recherche (ANR, French National Research Agency) -- 20-IADJ-0005-01 and the Deutsche Forschungsgemeinschaft (DFG, German Research Foundation) -- 442445488 under the trilateral ANR-DFG-JST call for project KEEPHA.
We would also like to thank the \textit{Lifeline} administrators for allowing us to use their data, explosion.ai for providing us with a researchers licence for Prodigy, and the anonymous reviewers for their detailed and helpful feedback.

\section{Bibliographical References}\label{reference}

\bibliographystyle{lrec2022-bib}
\bibliography{lrec_2022}

\section{Language Resource References}
\label{lr:ref}
\bibliographystylelanguageresource{lrec2022-bib}
\bibliographylanguageresource{languageresource}

\onecolumn
\section*{Appendix}

\input{appendix}

\begin{landscape}
\centering
\vspace*{\fill}
\begin{figure}[htb]
\resizebox{25cm}{!}{\input{exp_setting}}
\caption{The setup for the few-shot experiments:
(1 + 2) We fine-tune 10 XLM-R models on the English source language data (fine-tuning 1). 
(3) Then, we choose 5 seeds and create 5 train/dev sets, from which we sample the shots.
(4) We fine-tune (fine-tuning 2) each XLM-R model on each seed data, obtaining 10 XLM-R\_fine models for every seed.
(5) For every seed, each model is applied to the test set, and (6) we vote on the final results.
(7) We obtain 5 results, one for every seed (F1-scores etc).
(8)Those 5 results per setting are averaged.
}

\label{fig:exp_setup}
\end{figure}
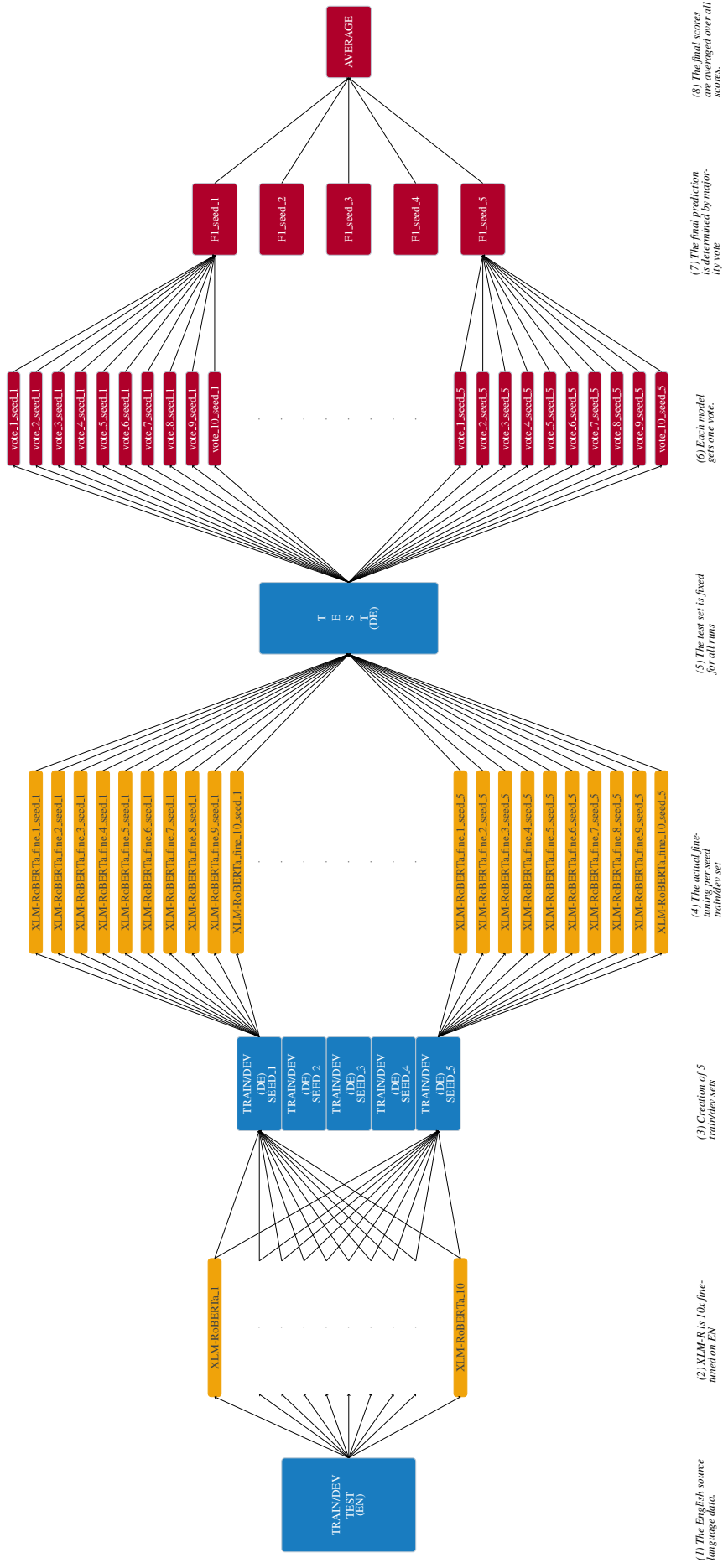
\vfill
\end{landscape}

\end{document}

%% file: results_table.tex
\begin{table*} \setlength{\tabcolsep}{2pt} 
\centering
\small
\begin{tabular}{@{}l*{12}{r}@{}}
\toprule
\multicolumn{1}{c}{} &
  \multicolumn{1}{c}{\textbf{method}} &
  \multicolumn{1}{c}{\textbf{data}} &
  \multicolumn{1}{c}{\textbf{P\_0}} &
  \multicolumn{1}{c}{\textbf{R\_0}} &
  \multicolumn{1}{c}{\textbf{F1\_0}} &
  \multicolumn{1}{c}{\textbf{P\_1}} &
  \multicolumn{1}{c}{\textbf{R\_1}} &
  \multicolumn{1}{c}{\textbf{F1\_1}} &
  \multicolumn{1}{c}{\textbf{P\_m}} &
  \multicolumn{1}{c}{\textbf{R\_m}} &
  \multicolumn{1}{c}{\textbf{F1\_m}} &
  \multicolumn{1}{c}{\textbf{AUC}} \\ \midrule \\
\rowcolor[HTML]{C0C0C0} 
SVM &
  full &
  all &
  98.73 &
  86.92 &
  92.5 &
  10.26 &
  57.14 &
  17.39 &
  54.49 &
  72.03 &
  54.92 &
  72.03 \\
SVM &
  per\_class &
  10 &
  \begin{tabular}[c]{@{}r@{}}99.34 \\ \footnotesize$\pm$ 1.01\end{tabular} &
  \begin{tabular}[c]{@{}r@{}}35.52 \\ \footnotesize$\pm$ 20.03\end{tabular} &
  \begin{tabular}[c]{@{}r@{}}49.48 \\ \footnotesize$\pm$ 23.39\end{tabular} &
  \begin{tabular}[c]{@{}r@{}}3.39 \\ \footnotesize$\pm$ 0.65\end{tabular} &
  \begin{tabular}[c]{@{}r@{}}85.71 \\ \footnotesize$\pm$ 24.28\end{tabular} &
  \begin{tabular}[c]{@{}r@{}}6.51 \\ \footnotesize$\pm$ 1.25\end{tabular} &
  \begin{tabular}[c]{@{}r@{}}51.36 \\ \footnotesize$\pm$ 0.7\end{tabular} &
  \begin{tabular}[c]{@{}r@{}}60.62 \\ \footnotesize$\pm$ 7.64\end{tabular} &
  \begin{tabular}[c]{@{}r@{}}27.99 \\ \footnotesize$\pm$ 11.88\end{tabular} &
  \begin{tabular}[c]{@{}r@{}}60.62 \\ \footnotesize$\pm$ 7.64\end{tabular} \\
\rowcolor[HTML]{C0C0C0} 
SVM &
  per\_class &
  40 &
  \textbf{\begin{tabular}[c]{@{}r@{}}99.90 \\ \footnotesize$\pm$ 0.22\end{tabular}} &
  \begin{tabular}[c]{@{}r@{}}22.24 \\ \footnotesize$\pm$ 4.73\end{tabular} &
  \begin{tabular}[c]{@{}r@{}}36.17 \\ \footnotesize$\pm$ 6.67\end{tabular} &
  \begin{tabular}[c]{@{}r@{}}3.23 \\ \footnotesize$\pm$ 0.17\end{tabular} &
  \textbf{\begin{tabular}[c]{@{}r@{}}99.05 \\ \footnotesize$\pm$ 2.13\end{tabular}} &
  \begin{tabular}[c]{@{}r@{}}6.26 \\ \footnotesize$\pm$ 0.31\end{tabular} &
  \begin{tabular}[c]{@{}r@{}}51.57 \\ \footnotesize$\pm$ 0.14\end{tabular} &
  \begin{tabular}[c]{@{}r@{}}60.64 \\ \footnotesize$\pm$ 2.23\end{tabular} &
  \begin{tabular}[c]{@{}r@{}}21.22 \\ \footnotesize$\pm$ 3.48\end{tabular} &
  \begin{tabular}[c]{@{}r@{}}60.64 \\ \footnotesize$\pm$ 2.23\end{tabular} \\
SVM &
  add\_neg &
  \begin{tabular}[c]{@{}r@{}}10 \\ + 200 neg\end{tabular} &
  \begin{tabular}[c]{@{}r@{}}98.27 \\ \footnotesize$\pm$ 0.17\end{tabular} &
  \begin{tabular}[c]{@{}r@{}}87.25 \\ \footnotesize$\pm$ 3.33\end{tabular} &
  \begin{tabular}[c]{@{}r@{}}92.4 \\ \footnotesize$\pm$ 1.77\end{tabular} &
  \begin{tabular}[c]{@{}r@{}}7.94 \\ \footnotesize$\pm$ 1.16\end{tabular} &
  \begin{tabular}[c]{@{}r@{}}40.95 \\ \footnotesize$\pm$ 7.97\end{tabular} &
  \begin{tabular}[c]{@{}r@{}}13.16 \\ \footnotesize$\pm$ 1.3\end{tabular} &
  \begin{tabular}[c]{@{}r@{}}53.11 \\ \footnotesize$\pm$ 0.56\end{tabular} &
  \begin{tabular}[c]{@{}r@{}}64.1 \\ \footnotesize$\pm$ 2.64\end{tabular} &
  \begin{tabular}[c]{@{}r@{}}52.78 \\ \footnotesize$\pm$ 1.38\end{tabular} &
  \begin{tabular}[c]{@{}r@{}}64.1 \\ \footnotesize$\pm$ 2.64\end{tabular} \\
\rowcolor[HTML]{C0C0C0} 
SVM &
  add\_neg &
  \begin{tabular}[c]{@{}r@{}}40 \\ + 400 neg\end{tabular} &
  \begin{tabular}[c]{@{}r@{}}98.98 \\ \footnotesize$\pm$ 0.33\end{tabular} &
  \begin{tabular}[c]{@{}r@{}}71.63 \\ \footnotesize$\pm$ 2.82\end{tabular} &
  \begin{tabular}[c]{@{}r@{}}83.08 \\ \footnotesize$\pm$ 1.8\end{tabular} &
  \begin{tabular}[c]{@{}r@{}}6.16 \\ \footnotesize$\pm$ 0.28\end{tabular} &
  \begin{tabular}[c]{@{}r@{}}71.43 \\ \footnotesize$\pm$ 10.1\end{tabular} &
  \begin{tabular}[c]{@{}r@{}}11.33 \\ \footnotesize$\pm$ 0.59\end{tabular} &
  \begin{tabular}[c]{@{}r@{}}52.57 \\ \footnotesize$\pm$ 0.3\end{tabular} &
  \begin{tabular}[c]{@{}r@{}}71.53 \\ \footnotesize$\pm$ 3.68\end{tabular} &
  \begin{tabular}[c]{@{}r@{}}47.21 \\ \footnotesize$\pm$ 0.68\end{tabular} &
  \begin{tabular}[c]{@{}r@{}}71.53 \\ \footnotesize$\pm$ 3.68\end{tabular} \\ \\ \midrule \\
BRB &
  Zero-shot &
  0 &
  97.55 &
  99.00 &
  98.27 &
  11.11 &
  4.76 &
  6.67 &
  54.33 &
  51.88 &
  52.47 &
  51.88 \\
\rowcolor[HTML]{C0C0C0} 
XLM-R &
  Zero-shot &
  0 &
  99.77 &
  54.42 &
  70.42 &
  5.18 &
  95.23 &
  9.82 &
  52.48 &
  74.83 &
  40.13 &
  74.83 \\ \\ \midrule \\
XLM-R &
  full &
  all &
  \begin{tabular}[c]{@{}r@{}}98.16 \\ \footnotesize$\pm$ 0.19\end{tabular} &
  \textbf{\begin{tabular}[c]{@{}r@{}}99.43 \\ \footnotesize$\pm$ 0.23\end{tabular}} &
  \textbf{\begin{tabular}[c]{@{}r@{}}98.79 \\ \footnotesize$\pm$ 0.07\end{tabular}} &
  \textbf{\begin{tabular}[c]{@{}r@{}}57.64\\ \footnotesize$\pm$ 7.14\end{tabular}} &
  \begin{tabular}[c]{@{}r@{}}28.57 \\ \footnotesize$\pm$ 7.53\end{tabular} &
  \textbf{\begin{tabular}[c]{@{}r@{}}37.52 \\ \footnotesize$\pm$ 6.65\end{tabular}} &
  \textbf{\begin{tabular}[c]{@{}r@{}}77.9 \\ \footnotesize$\pm$ 3.55\end{tabular}} &
  \begin{tabular}[c]{@{}r@{}}64.00\\ \footnotesize$\pm$ 3.68\end{tabular} &
  \textbf{\begin{tabular}[c]{@{}r@{}}68.15 \\ \footnotesize$\pm$ 3.33\end{tabular}} &
  \begin{tabular}[c]{@{}r@{}}64.00\\ \footnotesize$\pm$ 3.68\end{tabular} \\
\rowcolor[HTML]{C0C0C0} 
XLM-R &
  per\_class &
  10 &
  \begin{tabular}[c]{@{}r@{}}99.15 \\ \footnotesize$\pm$ 0.91\end{tabular} &
  \begin{tabular}[c]{@{}r@{}}66.45 \\ \footnotesize$\pm$ 9.91\end{tabular} &
  \begin{tabular}[c]{@{}r@{}}79.21 \\ \footnotesize$\pm$ 6.27\end{tabular} &
  \begin{tabular}[c]{@{}r@{}}5.24 \\ \footnotesize$\pm$ 1.25\end{tabular} &
  \begin{tabular}[c]{@{}r@{}}75.24 \\ \footnotesize$\pm$ 31.66\end{tabular} &
  \begin{tabular}[c]{@{}r@{}}9.75 \\ \footnotesize$\pm$ 2.55\end{tabular} &
  \begin{tabular}[c]{@{}r@{}}52.2 \\ \footnotesize$\pm$ 1.08\end{tabular} &
  \begin{tabular}[c]{@{}r@{}}70.84 \\ \footnotesize$\pm$ 10.88\end{tabular} &
  \begin{tabular}[c]{@{}r@{}}44.48 \\ \footnotesize$\pm$ 1.9\end{tabular} &
  \begin{tabular}[c]{@{}r@{}}70.84\\ \footnotesize$\pm$ 10.88\end{tabular} \\
XLM-R &
  per\_class &
  40 &
  \begin{tabular}[c]{@{}r@{}}99.71 \\ \footnotesize$\pm$ 0.13\end{tabular} &
  \begin{tabular}[c]{@{}r@{}}61.34 \\ \footnotesize$\pm$ 5.95\end{tabular} &
  \begin{tabular}[c]{@{}r@{}}75.82 \\ \footnotesize$\pm$ 4.69\end{tabular} &
  \begin{tabular}[c]{@{}r@{}}6.04 \\ \footnotesize$\pm$ 0.87\end{tabular} &
  \begin{tabular}[c]{@{}r@{}}93.33 \\ \footnotesize$\pm$ 2.61\end{tabular} &
  \begin{tabular}[c]{@{}r@{}}11.33 \\ \footnotesize$\pm$ 1.55\end{tabular} &
  \begin{tabular}[c]{@{}r@{}}52.87 \\ \footnotesize$\pm$ 0.48\end{tabular} &
  \begin{tabular}[c]{@{}r@{}}77.34 \\ \footnotesize$\pm$ 3.65\end{tabular} &
  \begin{tabular}[c]{@{}r@{}}43.58 \\ \footnotesize$\pm$ 3.11\end{tabular} &
  \begin{tabular}[c]{@{}r@{}}77.34 \\ \footnotesize$\pm$ 3.65\end{tabular} \\
\rowcolor[HTML]{C0C0C0} 
XLM-R &
  add\_neg &
  \begin{tabular}[c]{@{}r@{}}40 \\ + 100 neg\end{tabular} &
  \begin{tabular}[c]{@{}r@{}}98.22 \\ \footnotesize$\pm$ 0.73\end{tabular} &
  \begin{tabular}[c]{@{}r@{}}94.5 \\ \footnotesize$\pm$ 8.62\end{tabular} &
  \begin{tabular}[c]{@{}r@{}}96.12 \\ \footnotesize$\pm$ 4.46\end{tabular} &
  \begin{tabular}[c]{@{}r@{}}26.37 \\ \footnotesize$\pm$ 15.67\end{tabular} &
  \begin{tabular}[c]{@{}r@{}}32.38 \\ \footnotesize$\pm$ 30.38\end{tabular} &
  \begin{tabular}[c]{@{}r@{}}19.81 \\ \footnotesize$\pm$ 12.47\end{tabular} &
  \begin{tabular}[c]{@{}r@{}}62.29 \\ \footnotesize$\pm$ 7.67\end{tabular} &
  \begin{tabular}[c]{@{}r@{}}63.44 \\ \footnotesize$\pm$ 11.33\end{tabular} &
  \begin{tabular}[c]{@{}r@{}}57.97 \\ \footnotesize$\pm$ 6.94\end{tabular} &
  \begin{tabular}[c]{@{}r@{}}63.44 \\ \footnotesize$\pm$ 11.33\end{tabular} \\
XLM-R &
  add\_source &
  \begin{tabular}[c]{@{}r@{}}10 \\ + 100 neg \\ + 200 source\end{tabular} &
  \begin{tabular}[c]{@{}r@{}}99.39 \\ \footnotesize$\pm$ 0.27\end{tabular} &
  \begin{tabular}[c]{@{}r@{}}75.99 \\ \footnotesize$\pm$ 4.5\end{tabular} &
  \begin{tabular}[c]{@{}r@{}}86.07 \\ \footnotesize$\pm$ 2.84\end{tabular} &
  \begin{tabular}[c]{@{}r@{}}8.29 \\ \footnotesize$\pm$ 0.8\end{tabular} &
  \begin{tabular}[c]{@{}r@{}}81.9 \\ \footnotesize$\pm$ 8.52\end{tabular} &
  \begin{tabular}[c]{@{}r@{}}15.03 \\ \footnotesize$\pm$ 1.26\end{tabular} &
  \begin{tabular}[c]{@{}r@{}}53.84 \\ \footnotesize$\pm$ 0.36\end{tabular} &
  \textbf{\begin{tabular}[c]{@{}r@{}}78.95 \\ \footnotesize$\pm$ 2.67\end{tabular}} &
  \begin{tabular}[c]{@{}r@{}}50.55 \\ \footnotesize$\pm$ 1.99\end{tabular} &
  \textbf{\begin{tabular}[c]{@{}r@{}}78.95 \\ \footnotesize$\pm$ 2.67\end{tabular}} \\
\rowcolor[HTML]{C0C0C0} 
XLM-R &
  add\_source &
  \begin{tabular}[c]{@{}r@{}}40 \\ + 300 neg \\ + 300 source\end{tabular} &
  \begin{tabular}[c]{@{}r@{}}98.72 \\ \footnotesize$\pm$ 0.43\end{tabular} &
  \begin{tabular}[c]{@{}r@{}}90.91 \\ \footnotesize$\pm$ 4.82\end{tabular} &
  \begin{tabular}[c]{@{}r@{}}94.59 \\ \footnotesize$\pm$ 2.4\end{tabular} &
  \begin{tabular}[c]{@{}r@{}}15.84 \\ \footnotesize$\pm$ 6.53\end{tabular} &
  \begin{tabular}[c]{@{}r@{}}54.29 \\ \footnotesize$\pm$ 18.01\end{tabular} &
  \begin{tabular}[c]{@{}r@{}}22.55 \\ \footnotesize$\pm$ 3.42\end{tabular} &
  \begin{tabular}[c]{@{}r@{}}57.28 \\ \footnotesize$\pm$ 3.1\end{tabular} &
  \begin{tabular}[c]{@{}r@{}}72.6 \\ \footnotesize$\pm$ 6.7\end{tabular} &
  \begin{tabular}[c]{@{}r@{}}58.57 \\ \footnotesize$\pm$ 2.8\end{tabular} &
  \begin{tabular}[c]{@{}r@{}}72.6 \\ \footnotesize$\pm$ 6.7\end{tabular} \\ \\ \midrule \\
BRB &
  per\_class &
  10 &
  \begin{tabular}[c]{@{}r@{}}98.03 \\ \footnotesize$\pm$ 0.12\end{tabular} &
  \begin{tabular}[c]{@{}r@{}}75.54 \\ \footnotesize$\pm$ 5.29\end{tabular} &
  \begin{tabular}[c]{@{}r@{}}85.25 \\ \footnotesize$\pm$ 3.35\end{tabular} &
  \begin{tabular}[c]{@{}r@{}}4.38 \\ \footnotesize$\pm$ 0.7\end{tabular} &
  \begin{tabular}[c]{@{}r@{}}41.9 \\ \footnotesize$\pm$ 5.22\end{tabular} &
  \begin{tabular}[c]{@{}r@{}}7.91 \\ \footnotesize$\pm$ 1.12\end{tabular} &
  \begin{tabular}[c]{@{}r@{}}51.21 \\ \footnotesize$\pm$ 0.39\end{tabular} &
  \begin{tabular}[c]{@{}r@{}}58.72 \\ \footnotesize$\pm$ 1.98\end{tabular} &
  \begin{tabular}[c]{@{}r@{}}46.58 \\ \footnotesize$\pm$ 2.14\end{tabular} &
  \begin{tabular}[c]{@{}r@{}}58.72 \\ \footnotesize$\pm$ 1.98\end{tabular} \\
\rowcolor[HTML]{C0C0C0} 
BRB &
  per\_class &
  40 &
  \begin{tabular}[c]{@{}r@{}}99.14 \\ \footnotesize$\pm$ 0.23\end{tabular} &
  \begin{tabular}[c]{@{}r@{}}56.59 \\ \footnotesize$\pm$ 8.68\end{tabular} &
  \begin{tabular}[c]{@{}r@{}}71.72 \\ \footnotesize$\pm$ 7.39\end{tabular} &
  \begin{tabular}[c]{@{}r@{}}4.74 \\ \footnotesize$\pm$ 0.62\end{tabular} &
  \begin{tabular}[c]{@{}r@{}}80.95 \\ \footnotesize$\pm$ 6.73\end{tabular} &
  \begin{tabular}[c]{@{}r@{}}8.94 \\ \footnotesize$\pm$ 1.11\end{tabular} &
  \begin{tabular}[c]{@{}r@{}}51.94 \\ \footnotesize$\pm$ 0.35\end{tabular} &
  \begin{tabular}[c]{@{}r@{}}68.77 \\ \footnotesize$\pm$ 3.35\end{tabular} &
  \begin{tabular}[c]{@{}r@{}}40.33 \\ \footnotesize$\pm$ 4.2\end{tabular} &
  \begin{tabular}[c]{@{}r@{}}68.77 \\ \footnotesize$\pm$ 3.35\end{tabular} \\
BRB &
  add\_neg &
  \begin{tabular}[c]{@{}r@{}}40 \\ + 100 neg\end{tabular} &
  \begin{tabular}[c]{@{}r@{}}97.67 \\ \footnotesize$\pm$ 0.2\end{tabular} &
  \begin{tabular}[c]{@{}r@{}}99.00\\ \footnotesize$\pm$ 1.01\end{tabular} &
  \begin{tabular}[c]{@{}r@{}}98.33 \\ \footnotesize$\pm$ 0.4\end{tabular} &
  \begin{tabular}[c]{@{}r@{}}21.91 \\ \footnotesize$\pm$ 20.74\end{tabular} &
  \begin{tabular}[c]{@{}r@{}}9.52 \\ \footnotesize$\pm$ 8.69\end{tabular} &
  \begin{tabular}[c]{@{}r@{}}11.00\\ \footnotesize$\pm$ 8.6\end{tabular} &
  \begin{tabular}[c]{@{}r@{}}59.79 \\ \footnotesize$\pm$ 10.38\end{tabular} &
  \begin{tabular}[c]{@{}r@{}}54.26 \\ \footnotesize$\pm$ 3.86\end{tabular} &
  \begin{tabular}[c]{@{}r@{}}54.66 \\ \footnotesize$\pm$ 4.12\end{tabular} &
  \begin{tabular}[c]{@{}r@{}}54.26 \\ \footnotesize$\pm$ 3.86\end{tabular} \\
\rowcolor[HTML]{C0C0C0} 
BRB &
  add\_source &
  \begin{tabular}[c]{@{}r@{}}40 \\ + 100 neg \\ + 200 source\end{tabular} &
  \begin{tabular}[c]{@{}r@{}}97.94 \\ \footnotesize$\pm$ 0.11\end{tabular} &
  \begin{tabular}[c]{@{}r@{}}98.23 \\ \footnotesize$\pm$ 0.48\end{tabular} &
  \begin{tabular}[c]{@{}r@{}}98.08 \\ \footnotesize$\pm$ 0.23\end{tabular} &
  \begin{tabular}[c]{@{}r@{}}24.21 \\ \footnotesize$\pm$ 5.6\end{tabular} &
  \begin{tabular}[c]{@{}r@{}}20.95 \\ \footnotesize$\pm$ 4.26\end{tabular} &
  \begin{tabular}[c]{@{}r@{}}22.23 \\ \footnotesize$\pm$ 4.06\end{tabular} &
  \begin{tabular}[c]{@{}r@{}}61.07 \\ \footnotesize$\pm$ 2.83\end{tabular} &
  \begin{tabular}[c]{@{}r@{}}59.59 \\ \footnotesize$\pm$ 2.06\end{tabular} &
  \begin{tabular}[c]{@{}r@{}}60.16 \\ \footnotesize$\pm$ 2.08\end{tabular} &
  \begin{tabular}[c]{@{}r@{}}59.59 \\ \footnotesize$\pm$ 2.06\end{tabular} \\ \\ \bottomrule
\end{tabular}
\caption{Target language (German): results of the best runs for every scenario. We excluded those that scored an F1 of 0.0 for the positive class. BRB = BioRedditBERT, XLM-R = XLM-RoBERTa. \textbf{\_0} and \textbf{\_1} represent the negative and positive class, respectively. \textbf{P} is precision, \textbf{R} is recall and \textbf{F1} is F1-score, and \textbf{\_m} indicates the macro scores.}
\label{tab:final_results}
\end{table*}

%% file: appendix.tex
\begin{table}[h]
\centering
\begin{tabular}{@{}lr
>{\columncolor[HTML]{C0C0C0}}r 
>{\columncolor[HTML]{C0C0C0}}r 
>{\columncolor[HTML]{C0C0C0}}r rrr
>{\columncolor[HTML]{C0C0C0}}r 
>{\columncolor[HTML]{C0C0C0}}r 
>{\columncolor[HTML]{C0C0C0}}r r@{}}
\toprule
\multicolumn{1}{c}{\textbf{}} &
  \multicolumn{1}{c}{\textbf{}} &
  \multicolumn{3}{c}{\cellcolor[HTML]{C0C0C0}\textbf{Class 0}} &
  \multicolumn{3}{c}{\textbf{Class 1}} &
  \multicolumn{3}{c}{\cellcolor[HTML]{C0C0C0}\textbf{Macro Average}} &
  \multicolumn{1}{c}{\textbf{}} \\ \midrule
\multicolumn{1}{c}{\textbf{model}} &
  \multicolumn{1}{c}{\textbf{seed}} &
  \multicolumn{1}{c}{\textbf{P}} &
  \multicolumn{1}{c}{\textbf{R}} &
  \multicolumn{1}{c}{\textbf{F1}} &
  \multicolumn{1}{c}{\textbf{P}} &
  \multicolumn{1}{c}{\textbf{R}} &
  \multicolumn{1}{c}{\textbf{F1}} &
  \multicolumn{1}{c}{\textbf{P}} &
  \multicolumn{1}{c}{\textbf{R}} &
  \multicolumn{1}{c}{\textbf{F1}} &
  \multicolumn{1}{c}{\textbf{AUC}} \\
XLM\_R\_1  & 78   & 66.67 & 71.26 & 68.89 & 92.42 & 90.77 & 91.59 & 79.55 & 81.02 & 80.24 & 81.02 \\
XLM\_R\_2  & 99   & 68.57 & 55.17 & 61.15 & 88.95 & 93.45 & 91.15 & 78.76 & 74.31 & 76.15 & 74.31 \\
XLM\_R\_3  & 227  & 62.77 & 67.82 & 65.19 & 91.49 & 89.58 & 90.53 & 77.13 & 78.70 & 77.86 & 78.70 \\
XLM\_R\_4  & 409  & 66.25 & 60.92 & 63.47 & 90.09 & 91.96 & 91.02 & 78.17 & 76.44 & 77.24 & 76.44 \\
XLM\_R\_5  & 422  & 70.77 & 52.87 & 60.53 & 88.55 & 94.35 & 91.35 & 79.66 & 73.61 & 75.94 & 73.61 \\
XLM\_R\_6  & 482  & 64.89 & 70.11 & 67.40 & 92.10 & 90.18 & 91.13 & 78.50 & 80.15 & 79.27 & 80.15 \\
XLM\_R\_7  & 485  & 59.48 & 79.31 & 67.98 & 94.14 & 86.01 & 89.89 & 76.81 & 82.66 & 78.94 & 82.66 \\
XLM\_R\_8  & 841  & 61.22 & 68.97 & 64.86 & 91.69 & 88.69 & 90.17 & 76.46 & 78.83 & 77.52 & 78.83 \\
XLM\_R\_9  & 857  & 67.90 & 63.22 & 65.48 & 90.64 & 92.26 & 91.45 & 79.27 & 77.74 & 78.46 & 77.74 \\
XLM\_R\_10 & 910  & 71.43 & 63.22 & 67.07 & 90.75 & 93.45 & 92.08 & 81.09 & 78.34 & 79.58 & 78.34 \\
           & \textbf{mean} & 66.00 & 65.29 & 65.20 & 91.08 & 91.07 & 91.03 & 78.54 & 78.18 & 78.12 & 78.18 \\
           & \textbf{std}  & 3.94  & 7.89  & 2.81  & 1.67  & 2.55  & 0.67  & 1.45  & 2.82  & 1.44  & 2.82  \\ \bottomrule
\end{tabular}
\caption{Source language data (English): results for XLM-RoBERTa in precision, recall and F1 score per class and macro-averaged. The models have the same configuration and are trained and tested on the exact same data, but have a different seed for initialization.
Support for class 0: 87, support for class 1: 336}
\label{tab:english_results_xlm}
\end{table}

\begin{table*}[h]
\centering
\begin{tabular}{@{}lr
>{\columncolor[HTML]{C0C0C0}}r 
>{\columncolor[HTML]{C0C0C0}}r 
>{\columncolor[HTML]{C0C0C0}}r rrr
>{\columncolor[HTML]{C0C0C0}}r 
>{\columncolor[HTML]{C0C0C0}}r 
>{\columncolor[HTML]{C0C0C0}}r r@{}}
\toprule
\multicolumn{1}{c}{} &
  \multicolumn{1}{c}{\textbf{}} &
  \multicolumn{3}{c}{\cellcolor[HTML]{C0C0C0}\textbf{Class 0}} &
  \multicolumn{3}{c}{\textbf{Class 1}} &
  \multicolumn{3}{c}{\cellcolor[HTML]{C0C0C0}\textbf{Macro Average}} &
  \multicolumn{1}{c}{\textbf{}} \\ \midrule
\multicolumn{1}{c}{\textbf{model}} &
  \multicolumn{1}{c}{\textbf{seed}} &
  \multicolumn{1}{c}{\textbf{P}} &
  \multicolumn{1}{c}{\textbf{R}} &
  \multicolumn{1}{c}{\textbf{F1}} &
  \multicolumn{1}{c}{\textbf{P}} &
  \multicolumn{1}{c}{\textbf{R}} &
  \multicolumn{1}{c}{\textbf{F1}} &
  \multicolumn{1}{c}{\textbf{P}} &
  \multicolumn{1}{c}{\textbf{R}} &
  \multicolumn{1}{c}{\textbf{F1}} &
  \multicolumn{1}{c}{\textbf{AUC}} \\
BRB\_1    & 78               & 75.41 & 52.87 & 62.16 & 88.67 & 95.54 & 91.98 & 82.04 & 74.20 & 77.07 & 74.20 \\
BRB\_2    & 99               & 68.82 & 73.56 & 71.11 & 93.03 & 91.37 & 92.19 & 80.92 & 82.47 & 81.65 & 82.47 \\
BRB\_3    & 227              & 66.67 & 73.56 & 69.95 & 92.97 & 90.48 & 91.70 & 79.82 & 82.02 & 80.82 & 82.02 \\
BRB\_4    & 409              & 60.16 & 85.06 & 70.48 & 95.67 & 85.42 & 90.25 & 77.91 & 85.24 & 80.36 & 85.24 \\
BRB\_5    & 422              & 64.36 & 74.71 & 69.15 & 93.17 & 89.29 & 91.19 & 78.76 & 82.00 & 80.17 & 82.00 \\
BRB\_6    & 482              & 73.26 & 72.41 & 72.83 & 92.88 & 93.15 & 93.02 & 83.07 & 82.78 & 82.92 & 82.78 \\
BRB\_7    & 485              & 62.50 & 63.22 & 62.86 & 90.45 & 90.18 & 90.31 & 76.47 & 76.70 & 76.59 & 76.70 \\
BRB\_8    & 841              & 61.22 & 68.97 & 64.86 & 91.69 & 88.69 & 90.17 & 76.46 & 78.83 & 77.52 & 78.83 \\
BRB\_9    & 857              & 63.54 & 70.11 & 66.67 & 92.05 & 89.58 & 90.80 & 77.80 & 79.85 & 78.73 & 79.85 \\
BRB\_10   & 910              & 78.33 & 54.02 & 63.95 & 88.98 & 96.13 & 92.42 & 83.66 & 75.08 & 78.18 & 75.08 \\
\textbf{} & \textbf{mean}    & 67.43 & 68.85 & 67.40 & 91.96 & 90.98 & 91.40 & 79.69 & 79.92 & 79.40 & 79.92 \\
\textbf{} & \textbf{std dev} & 6.32  & 9.79  & 3.79  & 2.11  & 3.23  & 1.01  & 2.64  & 3.64  & 2.10  & 3.64  \\ \bottomrule
\end{tabular}
\caption{Source language data (English): results for BioRedditBERT in precision, recall and F1 score per class and macro-averaged. The models have the same configuration and are trained and tested on the exact same data, but have a different seed for initialization.
Support for class 0: 87, support for class 1: 336}
\label{tab:english_results_brb}
\end{table*}

\begin{table}[h]
\centering
\begin{tabular}{@{}llllll@{}}
\toprule
\textbf{model} & \textbf{data} & \textbf{learning rate} & \textbf{batch size} & \textbf{freeze} & \textbf{train sampler} \\ \midrule
XLM-R & English       & 0.00001056 & 7 & 1 & random   \\
BRB   & English       & 0.00001584 & 8 & 1 & random   \\
XLM-R & German (full) & 0.00001056 & 7 & 0 & weighted \\ \bottomrule
\end{tabular}
\caption{Specifications of the best models. The first and second lines correspond to the basis for the few-shot experiments where we trained 10 versions, the bottom one is XLM-RoBERTa again fine-tuned on the German full dataset. For the first two, a random sampler and freezing all layers except the classifier worked best, while not freezing any layers and using a weighted training sampler achieved the best performance for the third model.}
\label{tab:hyperparams}
\end{table}

%% file: exp_setting.tex
\begin{tikzpicture}
\node[draw=Gray,text width=4cm,minimum height=6cm, fill=L-lig, rounded corners, text centered, text=white] (en) at (-6,6) {\Large TRAIN/DEV\\TEST\\(EN)};
\node[text width=5cm] (comment_rob) at (-6,-10) {\Large\emph{(1) The English source language data.}};

\node[draw=Orange,fill=Orange,text=G-lig,text width=6cm,rounded corners, text centered] (rob1) at (2,12) {\Large XLM-RoBERTa\_1};

\node[minimum width=6cm] (rob2) at (2,10) {.};
\node[minimum width=6cm] (rob3) at (2,9) {.};
\node[minimum width=6cm] (rob4) at (2,8) {.};
\node[minimum width=6cm] (rob5) at (2,7) {.};
\node[minimum width=6cm] (rob6) at (2,6) {.};
\node[minimum width=6cm] (rob7) at (2,5) {.};
\node[minimum width=6cm] (rob8) at (2,4) {.};
\node[minimum width=6cm] (rob9) at (2,3) {.};

\node[draw=Orange,fill=Orange,text=G-lig,text width=6cm, rounded corners, text centered] (rob10) at (2,1) {\Large XLM-RoBERTa\_10};

\node[text width=5cm] (comment_rob) at (2,-10) {\Large \emph{ (2) XLM-R is 10x fine-tuned on EN}};

\node[draw=Gray,text width=4cm,minimum height=2cm, fill=L-lig, rounded corners, text centered, text=white] (traindev1) at (13,10) {\Large TRAIN/DEV\\(DE)\\SEED\_1};

\node[draw=Gray,text width=4cm,minimum height=2cm, fill=L-lig, rounded corners, text centered, text=white] (traindev2) at (13,8) {\Large TRAIN/DEV\\(DE)\\SEED\_2};

\node[draw=Gray,text width=4cm,minimum height=2cm, fill=L-lig, rounded corners, text centered, text=white] (traindev3) at (13,6) {\Large TRAIN/DEV\\(DE)\\SEED\_3};

\node[draw=Gray,text width=4cm,minimum height=2cm, fill=L-lig, rounded corners, text centered, text=white] (traindev4) at (13,4) {\Large TRAIN/DEV\\(DE)\\SEED\_4};

\node[draw=Gray,text width=4cm,minimum height=2cm, fill=L-lig, rounded corners, text centered, text=white] (traindev5) at (13,2) {\Large TRAIN/DEV\\(DE)\\SEED\_5};

\node[text width=5cm] (comment_traindev) at (13,-10) {\Large \emph{(3) Creation of 5 train/dev sets}};


\node[draw=Orange,fill=Orange,text=G-lig,text width=8cm,rounded corners, text centered] (robfineseed11) at (23,20) {\Large XLM-RoBERTa\_fine\_1\_seed\_1};
\node[draw=Orange,fill=Orange,text=G-lig,text width=8cm,rounded corners, text centered] (robfineseed12) at (23,19) {\Large XLM-RoBERTa\_fine\_2\_seed\_1};
\node[draw=Orange,fill=Orange,text=G-lig,text width=8cm,rounded corners, text centered] (robfineseed13) at (23,18) {\Large XLM-RoBERTa\_fine\_3\_seed\_1};
\node[draw=Orange,fill=Orange,text=G-lig,text width=8cm,rounded corners, text centered] (robfineseed14) at (23,17) {\Large XLM-RoBERTa\_fine\_4\_seed\_1};
\node[draw=Orange,fill=Orange,text=G-lig,text width=8cm,rounded corners, text centered] (robfineseed15) at (23,16) {\Large XLM-RoBERTa\_fine\_5\_seed\_1};
\node[draw=Orange,fill=Orange,text=G-lig,text width=8cm,rounded corners, text centered] (robfineseed16) at (23,15) {\Large XLM-RoBERTa\_fine\_6\_seed\_1};
\node[draw=Orange,fill=Orange,text=G-lig,text width=8cm,rounded corners, text centered] (robfineseed17) at (23,14) {\Large XLM-RoBERTa\_fine\_7\_seed\_1};
\node[draw=Orange,fill=Orange,text=G-lig,text width=8cm,rounded corners, text centered] (robfineseed18) at (23,13) {\Large XLM-RoBERTa\_fine\_8\_seed\_1};
\node[draw=Orange,fill=Orange,text=G-lig,text width=8cm,rounded corners, text centered] (robfineseed19) at (23,12) {\Large XLM-RoBERTa\_fine\_9\_seed\_1};
\node[draw=Orange,fill=Orange,text=G-lig,text width=8cm,rounded corners, text centered] (robfineseed110) at (23,11) {\Large XLM-RoBERTa\_fine\_10\_seed\_1};

\node[minimum width=8cm] (robfine2) at (23,10) {.};
\node[minimum width=8cm] (robfine3) at (23,9) {.};
\node[minimum width=8cm] (robfine4) at (23,8) {.};
\node[minimum width=8cm] (robfine5) at (23,7) {.};
\node[minimum width=8cm] (robfine6) at (23,6) {.};
\node[minimum width=8cm] (robfine7) at (23,5) {.};
\node[minimum width=8cm] (robfine8) at (23,4) {.};
\node[minimum width=8cm] (robfine9) at (23,3) {.};

\node[draw=Orange,fill=Orange,text=G-lig,text width=8cm, rounded corners, text centered] (robfineseed51) at (23,1) {\Large XLM-RoBERTa\_fine\_1\_seed\_5};

\node[draw=Orange,fill=Orange,text=G-lig,text width=8cm, rounded corners, text centered] (robfineseed52) at (23,0) {\Large XLM-RoBERTa\_fine\_2\_seed\_5};

\node[draw=Orange,fill=Orange,text=G-lig,text width=8cm, rounded corners, text centered] (robfineseed53) at (23,-1) {\Large XLM-RoBERTa\_fine\_3\_seed\_5};
\node[draw=Orange,fill=Orange,text=G-lig,text width=8cm, rounded corners, text centered] (robfineseed54) at (23,-2) {\Large XLM-RoBERTa\_fine\_4\_seed\_5};
\node[draw=Orange,fill=Orange,text=G-lig,text width=8cm, rounded corners, text centered] (robfineseed55) at (23,-3) {\Large XLM-RoBERTa\_fine\_5\_seed\_5};
\node[draw=Orange,fill=Orange,text=G-lig,text width=8cm, rounded corners, text centered] (robfineseed56) at (23,-4) {\Large XLM-RoBERTa\_fine\_6\_seed\_5};
\node[draw=Orange,fill=Orange,text=G-lig,text width=8cm, rounded corners, text centered] (robfineseed57) at (23,-5) {\Large XLM-RoBERTa\_fine\_7\_seed\_5};
\node[draw=Orange,fill=Orange,text=G-lig,text width=8cm, rounded corners, text centered] (robfineseed58) at (23,-6) {\Large XLM-RoBERTa\_fine\_8\_seed\_5};
\node[draw=Orange,fill=Orange,text=G-lig,text width=8cm, rounded corners, text centered] (robfineseed59) at (23,-7) {\Large XLM-RoBERTa\_fine\_9\_seed\_5};
\node[draw=Orange,fill=Orange,text=G-lig,text width=8cm, rounded corners, text centered] (robfineseed510) at (23,-8) {\Large XLM-RoBERTa\_fine\_10\_seed\_5};

\node[text width=5cm] (comment_robfine) at (23,-10) {\Large \emph{(4) The actual fine-tuning per seed train/dev set}};


\node[draw=Gray,fill=L-lig,minimum height=8cm, text=white,text width=3cm,rounded corners, text centered] (test) at (34,6) {\Large T\\E\\S\\T\\(DE)};
\node[text width=5cm] (comment_test) at (34,-10) {\Large{\emph{(5) The test set is fixed for all runs}}};

\node[draw=Gray,fill=DarkRed,text=white,text width=4cm,rounded corners, text centered] (voteseed11) at (43,21) {\Large vote\_1\_seed\_1};
\node[draw=Gray,fill=DarkRed,text=white,text width=4cm,rounded corners, text centered] (voteseed12) at (43,20) {\Large vote\_2\_seed\_1};
\node[draw=Gray,fill=DarkRed,text=white,text width=4cm,rounded corners, text centered] (voteseed13) at (43,19) {\Large vote\_3\_seed\_1};
\node[draw=Gray,fill=DarkRed,text=white,text width=4cm,rounded corners, text centered] (voteseed14) at (43,18) {\Large vote\_4\_seed\_1};
\node[draw=Gray,fill=DarkRed,text=white,text width=4cm,rounded corners, text centered] (voteseed15) at (43,17) {\Large vote\_5\_seed\_1};
\node[draw=Gray,fill=DarkRed,text=white,text width=4cm,rounded corners, text centered] (voteseed16) at (43,16) {\Large vote\_6\_seed\_1};
\node[draw=Gray,fill=DarkRed,text=white,text width=4cm,rounded corners, text centered] (voteseed17) at (43,15) {\Large vote\_7\_seed\_1};
\node[draw=Gray,fill=DarkRed,text=white,text width=4cm,rounded corners, text centered] (voteseed18) at (43,14) {\Large vote\_8\_seed\_1};
\node[draw=Gray,fill=DarkRed,text=white,text width=4cm,rounded corners, text centered] (voteseed19) at (43,13) {\Large vote\_9\_seed\_1};
\node[draw=Gray,fill=DarkRed,text=white,text width=4cm,rounded corners, text centered] (voteseed110) at (43,12)  {\Large vote\_10\_seed\_1};

\node[minimum width=4cm] (vote2) at (43,10) {.};
\node[minimum width=4cm] (vote3) at (43,9) {.};
\node[minimum width=4cm] (vote4) at (43,8) {.};
\node[minimum width=4cm] (vote5) at (43,7) {.};
\node[minimum width=4cm] (vote6) at (43,6) {.};
\node[minimum width=4cm] (vote7) at (43,5) {.};
\node[minimum width=4cm] (vote8) at (43,4) {.};
\node[minimum width=4cm] (vote9) at (43,3) {.};

\node[draw=Gray,fill=DarkRed,text=white,text width=4cm,rounded corners, text centered] (voteseed51) at (43,1) {\Large vote\_1\_seed\_5};
\node[draw=Gray,fill=DarkRed,text=white,text width=4cm,rounded corners, text centered] (voteseed52) at (43,0) {\Large vote\_2\_seed\_5};
\node[draw=Gray,fill=DarkRed,text=white,text width=4cm,rounded corners, text centered] (voteseed53) at (43,-1) {\Large vote\_3\_seed\_5};
\node[draw=Gray,fill=DarkRed,text=white,text width=4cm,rounded corners, text centered] (voteseed54) at (43,-2) {\Large vote\_4\_seed\_5};
\node[draw=Gray,fill=DarkRed,text=white,text width=4cm,rounded corners, text centered] (voteseed55) at (43,-3) {\Large vote\_5\_seed\_5};
\node[draw=Gray,fill=DarkRed,text=white,text width=4cm,rounded corners, text centered] (voteseed56) at (43,-4) {\Large vote\_6\_seed\_5};
\node[draw=Gray,fill=DarkRed,text=white,text width=4cm,rounded corners, text centered] (voteseed57) at (43,-5) {\Large vote\_7\_seed\_5};
\node[draw=Gray,fill=DarkRed,text=white,text width=4cm,rounded corners, text centered] (voteseed58) at (43,-6) {\Large vote\_8\_seed\_5};
\node[draw=Gray,fill=DarkRed,text=white,text width=4cm,rounded corners, text centered] (voteseed59) at (43,-7) {\Large vote\_9\_seed\_5};
\node[draw=Gray,fill=DarkRed,text=white,text width=4cm,rounded corners, text centered] (voteseed510) at (43,-8) {\Large vote\_10\_seed\_5};

\node[text width=4cm] (comment_vote) at (43,-10) {\Large{\emph{(6) Each model gets one vote.}}};

\node[draw=Gray,fill=DarkRed,text=white,text width=3cm,minimum height=2cm,rounded corners, text centered] (result1) at (52,12) {\Large F1\_seed\_1};

\node[draw=Gray,fill=DarkRed,text=white,text width=3cm,minimum height=2cm,rounded corners, text centered] (result2) at (52,9) {\Large F1\_seed\_2};
\node[draw=Gray,fill=DarkRed,text=white,text width=3cm,minimum height=2cm,rounded corners, text centered] (result3) at (52,6) {\Large F1\_seed\_3};
\node[draw=Gray,fill=DarkRed,text=white,text width=3cm,minimum height=2cm,rounded corners, text centered] (result4) at (52,3) {\Large F1\_seed\_4};
\node[draw=Gray,fill=DarkRed,text=white,text width=3cm,minimum height=2cm,rounded corners, text centered] (result5) at (52,0) {\Large F1\_seed\_5};

\node[text width=5cm] (comment_result) at (52,-10) {\Large{\emph{(7) The final prediction is determined by majority vote}}};

\node[draw=Gray,fill=DarkRed,text=white,text width=3cm,minimum height=2cm,rounded corners, text centered] (avrg) at (60,6) {\Large AVERAGE};
\node[text width=5cm] (comment_result) at (60,-10) {\Large{\emph{(8) The final scores are averaged over all scores.}}};

\foreach \dest in {1,...,10}
    \draw[->] (en.east) to (rob\dest.west);

\foreach \source in {1,...,10}
    \draw[->] (rob\source.east) to (traindev1.west);

\foreach \source in {1,...,10}
    \draw[->] (rob\source.east) to (traindev5.west);

\foreach \dest in {1,...,10}
    \draw[->] (traindev1.east) to (robfineseed1\dest.west);

\foreach \dest in {1,...,10}
    \draw[->] (traindev5.east) to (robfineseed5\dest.west);


\foreach \source in {1,...,10}
    \draw[->] (robfineseed1\source.east) to (test.west);

\foreach \source in {1,...,10}
    \draw[->] (robfineseed5\source.east) to (test.west);
    
\foreach \dest in {1,...,10}
    \draw[->] (test.east) to (voteseed1\dest.west);

\foreach \dest in {1,...,10}
    \draw[->] (test.east) to (voteseed5\dest.west);

\foreach \source in {1,...,10}
    \draw[->] (voteseed1\source.east) to (result1.west);

\foreach \source in {1,...,10}
    \draw[->] (voteseed5\source.east) to (result5.west);

\foreach \source in {1,...,5}
    \draw[->] (result\source.east) to (avrg.west);

\end{tikzpicture}